\documentclass[letterpaper, 10 pt, conference]{ieeeconf}  

\IEEEoverridecommandlockouts                              

\title{\LARGE \bf
Deep Person Detection in 2D Range Data
}

\author{Lucas Beyer$^{1}$, Alexander Hermans$^{1}$, Timm Linder$^{2}$, Kai O. Arras$^{3}$ and Bastian Leibe$^{1}$
\thanks{$^{1}$Lucas Beyer, Alexander Hermans and Bastian Leibe are with the Visual Computing Institute, RWTH Aachen University.
        {\tt\small last@vision.rwth-aachen.de}}%
\thanks{$^{2}$Timm Linder is with Robert Bosch GmbH, Corporate Research, Stuttgart, Germany and with the University of Freiburg, Germany.
        {\tt\small first.last@de.bosch.com}}%
    \thanks{$^{3}$Kai O. Arras is with Robert Bosch GmbH, Corporate Research.}
}

\usepackage{makeidx}
\usepackage{graphicx}
\usepackage{subcaption}
\usepackage{amsmath,amssymb} 
\usepackage{color}
\usepackage{hyperref}
\usepackage{xspace}
\usepackage{tabularx}
\usepackage{tikz}
\usetikzlibrary{calc,shapes,arrows}
\usepackage{pgfplots}
\usepackage{siunitx}
\usepackage{booktabs}
\usepackage{csvsimple}

\newcommand{\PAR}[1]{\vskip4pt \noindent {\bf #1~}}  

\newcommand{\rf}[1]{{\protect\NoHyper\ref{#1}\protect\endNoHyper}}
\newcommand{\forceref}[1]{\raisebox{2pt}{\tikz{\draw[#1, line width=0.9pt](0,0) -- (5mm,0);}}}

\definecolor{wd_c}{HTML}{E24A33}
\definecolor{wc_c}{HTML}{348ABD}
\definecolor{wa_c}{HTML}{D175F0}
\definecolor{wp_c}{HTML}{8EBA42}
\definecolor{awesome_orange}{HTML}{F9A91F}
\definecolor{good_gray}{HTML}{777777}
\definecolor{deep_pink}{HTML}{FF69B4}
\definecolor{shit_brown}{HTML}{8B4513}
\definecolor{random_blue}{HTML}{7DE2F0}
\definecolor{evil_green}{HTML}{366D32}

\pgfplotsset{
    precrec/.style={
        inner sep=0pt,outer sep=0pt,
        ylabel style={font=\scriptsize,yshift=-18pt},
        xlabel style={font=\scriptsize,yshift=6pt},
        width={1.1\linewidth},
        height={1.1\linewidth},
        yticklabel style = {font=\scriptsize,xshift=-0.3ex},
        xticklabel style = {font=\scriptsize,yshift=-0.3ex},
        legend image post style={line width =1.5pt},
        tick label style={/pgf/number format/assume math mode=true},
        tick align=outside,
        major tick length=2pt,
        every tick/.style={black, thin},
        tick pos=left,
        axis line style={draw=none},
        xtick={0,20,...,100},
        ytick={0,20,...,100},
        xlabel={Recall [\%]},
        ylabel={Precision [\%]},
        xmin=-2,
        xmax=102,
        ymin=-2,
        ymax=102,
        axis background/.style={fill=black!11!white},
        grid=both,
        grid style={white}
    }
}

\pgfplotsset{legend image code/.code={%
                \draw[mark repeat=2,mark phase=2]
                plot coordinates {
                (0cm,0cm)
                (0.3cm,0cm)        
                (0.6cm,0cm)         
                };%
            }
}

\newenvironment{customlegend}[1][]{%
    \begingroup
    \csname pgfplots@init@cleared@structures\endcsname
    \pgfplotsset{#1}%
}{%
    \csname pgfplots@createlegend\endcsname
    \endgroup
}%
\def\addlegendimage{\csname pgfplots@addlegendimage\endcsname}

\newcommand{\sampleselect}[1]{data/pr_curves/#1_fast.csv}
\newcommand{\precrec}[3]{\addplot [color=#2, #3] table [y=prec, x=rec, col sep=comma] {\sampleselect{#1}};}
\newcommand{\meta}[2]{\csvreader[column count=3]{data/pr_curves/#1_meta.csv}{1=\metaauc, 2=\metapeakfone, 3=\metaeer}{#2}}
\newcommand{\eer}[1]{\meta{#1}{\num[detect-all,round-mode=places,round-precision=1]{\metaeer}}}
\newcommand{\peakfone}[1]{\meta{#1}{\num[detect-all,round-mode=places,round-precision=1]{\metapeakfone}}}
\newcommand{\auc}[1]{\meta{#1}{\num[detect-all,round-mode=places,round-precision=1]{\metaauc}}}
\newcommand{\aucpeakfoneeer}[1]{~\auc{#1}~~~~~~~\peakfone{#1}~~~~~~~~\eer{#1}}
\newcommand{\aucpeakfoneeerbbb}[1]{~\textbf{\auc{#1}}~~~~~~~\textbf{\peakfone{#1}}~~~~~~~~\textbf{\eer{#1}}}
\newcommand{\aucpeakfoneeeriii}[1]{~\textit{\auc{#1}}~~~~~~~\textit{\peakfone{#1}}~~~~~~~~\textit{\eer{#1}}}
\newcommand{\bl}{\addlegendimage{white, only marks}}

\makeatletter
\DeclareRobustCommand\onedot{\futurelet\@let@token\@onedot}
\def\@onedot{\ifx\@let@token.\else.\null\fi\xspace}

\def\eg{\emph{e.g}\onedot} 
\def\ie{\emph{i.e}\onedot} 
\def\cf{\emph{c.f}\onedot} 
\def\etc{\emph{etc}\onedot} 
 
\def\etal{\emph{et al}\onedot}
\makeatother

\begin{document}
    \maketitle
    \thispagestyle{empty}
    \pagestyle{empty}

    \begin{abstract}
    Detecting humans is a key skill for mobile robots and intelligent vehicles in a large variety of applications.
    While the problem is well studied for certain sensory modalities such as image data, few works exist that address this detection task using 2D range data.
    However, a widespread sensory setup for many mobile robots in service and domestic applications contains a horizontally mounted 2D laser scanner.
    Detecting people from 2D range data is challenging due to the speed and dynamics of human leg motion and the high levels of occlusion and self-occlusion particularly in crowds of people.
    While previous approaches mostly relied on handcrafted features, we recently developed the deep learning based wheelchair and walker detector DROW.
    In this paper, we show the generalization to people, including small modifications that significantly boost DROW's performance.
    Additionally, by providing a small, fully online temporal window in our network, we further boost our score.
    We extend the DROW dataset with person annotations, making this the largest dataset of person annotations in 2D range data, recorded during several days in a real-world environment with high diversity.
    Extensive experiments with three current baseline methods indicate it is a challenging dataset, on which our improved DROW detector beats the current state-of-the-art.
    \end{abstract}

    \section{INTRODUCTION}
    Robots that share spaces with people require the ability to detect, track and analyze humans for a variety of reasons including safety, user interaction, or efficient navigation through crowds~\cite{Triebel16FSR,Hawes17RAM}.
    Self-driving cars~\cite{Geiger12CVPR} or last-mile delivery robots, for example, need to recognize pedestrians and particularly also vulnerable road users such as elderly people, children or people in wheelchairs.
    This recognition task is well understood for image data~\cite{Redmon17CVPR,Lin17ICCV}, sometimes even specifically geared toward robust person detection~\cite{Zhang2017CVPR}, and with systems for pedestrian detection being commercially available for automotive use-cases.
    For other sensory modalities such as RGB-D or 2D/3D range data, often needed in robotics, a growing number of detection methods and datasets exist.

    Many interesting approaches have been developed for detection in volumetric data, such as 3D LiDAR, which can largely be grouped into two types.
    The first type uses ordered projections of the point cloud data~\cite{Maturana15IROS,Ondruska16RSSWorkshop}.
    The second, more recent type proposes novel neural network architectures which can consume unordered point cloud data and reason on them, \eg PointNet~\cite{Qi17CVPR} and VoxelNet~\cite{Zhou17Arxiv}.

    Despite the advent of increasingly performant and affordable 3D/RGB-D sensors, 2D horizontally mounted laser scanners are still a common part of the sensory setup on a large number of mobile service robots in human populated spaces, making them an interesting sensor for person detection.
    They are also one of the few certifiable sensors to ensure human safety \eg for transportation platforms in intralogistics or hospitals, have large fields of view, high accuracy, and high levels of robustness in wide ranges of conditions.
    In contrast, on-board cameras have several drawbacks in many robotic use-cases including limited field of view, ambiguous scale, and likely detection failures when subjects are close to the sensor, an important case for human-robot interaction.

    Existing 2D range data based person detectors can roughly be grouped into two types.
    The first type consists of jump-distance based clustering of the measured laser points, computation of many hand-crafted features on these clusters, and use of a machine learning approach to classify the cluster as either person or background using said features~\cite{Arras07ICRA,Spinello08ICRA,Weinrich14ROMAN}.
    The second type does the same thing, although for legs as opposed to persons, by tracking legs over time, and defining persons as pairs of legs which satisfy some hand-designed heuristics~\cite{Pantofaru10ROS,Leigh15ICRA,Weinrich14ROMAN}.

    The latter type typically performs better on benchmark datasets and toy deployments, but the ``two leg assumption'' comes with some severe inherent drawbacks:
    1) It does not readily extend to other classes, \eg wheelchairs.
    2) It only works when the legs are visible, which is neither true in many real-world scenarios due to people wearing coats, dresses, hijabs, saris, \etc, nor in many professional scenarios due to work uniforms such as lab coats.
    This is a perfect example of why it is desirable to collect a diverse dataset and directly learn from data instead of  hand-crafting large parts of a system.
    Learning from 2D range data is difficult as evidenced by the relatively poorer performance of the jump distance-based methods when compared to the leg tracking methods.
    One big difference is that the latter often make use of rich temporal information.
    While deep learning has made impressive progress, also for detection on range data, so far our DROW~\cite{Beyer16RAL} detector is the only deep learning based detector for 2D range data.
    DROW obtained state-of-the-art detection results for walking aids and has performed well in multiple real-world deployments.

    In this paper we extend DROW to person detection.
    Given the multi-class nature of DROW, this is as straight forward as adding an additional class.
    However, given that persons in 2D range data are usually covered by few points, person detection proved to be a challenging task for the DROW detector too.

    For every point in a laser scan the DROW detector predicts a class and possibly a vote to the closest object center.
    A final post-processing step then converts votes to actual predictions.
    We improve this post-processing step, which now allows us to predict confidences for each detection and improves the overall performance.
    Additionally, we slightly increase our convolutional neural network (CNN) size, further boosting detection scores.

    Orthogonal to these extensions, we experiment with a fully online temporal integration of a few previous scans and show that this results in a significant performance boost for person detection scores.
    Unlike a tracker, we do not perform any data association, but rather show how DROW's input pre-processing can be extended to include temporal context to be used by the network.
    We show how several fusion strategies in the network compare and obtain state-of-the-art results using our extended DROW detector.

    To facilitate all our experiments, we introduce a new dataset of people annotations for our original DROW dataset.
    We hand annotate persons for every frame which we previously annotated with wheelchair and walker center points.
    To the best of our knowledge, this is the largest dataset for person detection in 2D range data recorded in real-world scenarios with over 10 hours of laser scan recordings.

    We thoroughly evaluate top performing existing detectors on this dataset, including retraining and tuning hyperparameters on our validation set.
    Since person detection is arguably a bigger field than walking aid detection in the robotics community, with the addition of person annotations, the DROW dataset is now by far more interesting and relevant.
    Our final results and the complete DROW dataset will be shared with the community to foster further research in this area.

    To summarize, our key contributions are as follows:
    \begin{enumerate}
    \item We extend the DROW detector to persons and propose several improvements boosting overall performance.
    \item A temporal fusion extension to integrate past frames into the detector, achieving state-of-the-art detection results for persons and their walking aids.
    \item A new large-scale ``in the wild'' indoor 2D laser dataset for person detection.
    \item A thorough evaluation of current state-of-the-art person detectors on our dataset.
    \end{enumerate}







    \newcommand{\pb}[3]{%
        \hspace*{-7pt}
        \begin{tikzpicture}[baseline={($ (current bounding box.north) - (0,8pt) $)}]
            \begin{axis}[axis lines=none,
                         ticks=none,
                         xticklabel=\empty,
                         yticklabel=\empty,
                         height=2.3cm,
                         width=3.15cm,
                         ybar, ybar interval]
                \addplot [color=#1, fill=#1, fill opacity=0.4] table [x=dist, y=#2_#3] {data/dataset_statistics/all.dat};
            \end{axis}
        \end{tikzpicture}%
            \vspace*{1pt}
    }
    \newcommand{\pbagnostic}[2]{%
        \hspace*{-7pt}
        \begin{tikzpicture}[baseline={($ (current bounding box.north) - (0,8pt) $)}]
            \begin{axis}[axis lines=none,
                         ticks=none,
                         xticklabel=\empty,
                         yticklabel=\empty,
                         height=2.3cm,
                         width=3.15cm,
                         ybar, ybar interval]
                \addplot [color=#1, fill=#1, fill opacity=0.4] table [x=dist, y expr=\thisrow{wc_#2} + \thisrow{wa_#2} + \thisrow{wp_#2}] {data/dataset_statistics/all.dat};
            \end{axis}
        \end{tikzpicture}%
            \vspace*{1pt}
    }
    \newcommand{\lc}[1]{\multicolumn{1}{c}{#1}}
    \newcolumntype{Y}{>{\raggedleft\arraybackslash}X}
    \begin{table}[t]

    \caption{DROW Dataset overview}
    \label{table:dataset}
    \begin{tabularx}{\linewidth}{p{1.9cm}YYYY}
        \toprule
                             & \lc{Train}     & \lc{Validation} & \lc{Test}     & \lc{Total}     \\\midrule
         Sequences           & \lc{78}        & \lc{30}         & \lc{5}        & \lc{113}       \\
         Scans               & 341\,138\,\,\, & 74\,744\,\,\,   & 48\,131\,\,\, & 464\,013\,\,\, \\
         Annotated Scans     &  17\,665\,\,\, &  3\,919\,\,\,   &  2\,428\,\,\, &  24\,012\,\,\, \\
         Wheelchairs         &  14\,455\,\,\, &  5\,595\,\,\,   &  1\,970\,\,\, &  22\,020\,\,\, \\
         Walkers             &   2\,047\,\,\, &     219\,\,\,   &     581\,\,\, &   2\,847\,\,\, \\
         Persons             &  23\,906\,\,\, &  3\,102\,\,\,   &  1\,976\,\,\, &  28\,984\,\,\, \\
         Wheelchairs\newline by distance & \pb{wc_c}{wc}{train} & \pb{wc_c}{wc}{valid} & \pb{wc_c}{wc}{test} & \pb{wc_c}{wc}{all} \\
         Walkers    \newline by distance & \pb{wa_c}{wa}{train} & \pb{wa_c}{wa}{valid} & \pb{wa_c}{wa}{test} & \pb{wa_c}{wa}{all} \\
         Person     \newline by distance & \pb{wp_c}{wp}{train} & \pb{wp_c}{wp}{valid} & \pb{wp_c}{wp}{test} & \pb{wp_c}{wp}{all} \\
         All        \newline by distance & \pbagnostic{wd_c}{train} & \pbagnostic{wd_c}{valid} & \pbagnostic{wd_c}{test} & \pbagnostic{wd_c}{all} \\
        \bottomrule
    \end{tabularx}
    \end{table}

    \section{THE DROW DATASET}
    We introduce a large new set of person annotations for the existing DROW dataset~\cite{Beyer16RAL}, which was recorded with a SICK S300 laser scanner at 12.5 Hz and an angular resolution of 1/2 degree at 37 cm above ground.
    To facilitate temporal approaches, we used the same annotation strategy as before:
    We divide the dataset into batches of 100 frames, annotating every fourth of these batches, while annotating every fifth frame within a batch.
    This results in 5\% of all frames being annotated, stretched over long time-frames, allowing for interesting temporal approaches (see Section~\ref{sec:temporal-integration}).
    We annotated exactly those frames which were previously annotated with wheelchairs and walkers, annotating persons that were walking on their own, as well as those that were pushing a walking aid.
    This means that a person pushing a walker results in two annotations: the person and the walker, while a person sitting in a wheelchair is only labeled as a wheelchair.
    We also considered persons sitting on normal chairs as persons, since these have a clearly different pattern in the data.

    Table~\ref{table:dataset} shows an overview of the resulting dataset.
    All numbers given in the table refer to individual annotations and not to instances, and the bar plots show their distribution over the distance.
    Each bar represents a \SI{1}{\m} slice in the distance (15 bars for up to \SI{15}{\m}).

    It should be noted that persons were significantly harder to annotate.
    By scrolling through a few frames and looking at the RGB images of the camera, it was fairly easy to find walking aid patterns in the scans.
    However, to annotate persons it was often necessary to focus on a single blob and ``track'' the pattern throughout a full batch of 100 frames in order to reliably tell which laser points represent a person.
    This clearly suggests that single frame detection of persons is a difficult task.


    Compared to the few publicly available 2D laser range datasets for person detection at leg height,
    ours is to our best knowledge the largest in number of raw and annotated scans, as well as covered duration and therefore much better suited for the investigation of deep-learning methods.
    While~\cite{Leigh15ICRA} introduced a dataset for person tracking in 2D laser at leg height, their indoor multi-person dataset is limited to three sequences of 12 minutes total duration, significantly smaller than ours.    
    Furthermore, due to the static background in the positive training data, their dataset is less suited for deep learning approaches which will easily overfit.

    %
    %
    %

    The home and reha datasets of~\cite{Weinrich14ROMAN} were recorded with a similar sensor setup and in a similar scenario as ours.
    While the number of annotated scans is similar, our raw scan count is larger by an order of magnitude and covers a significantly longer period of time.
    Our average person count per annotated frame is twice as high, suggesting that our dataset is more challenging.



    The Freiburg Main Station and City Center datasets~\cite{Luber11IJRR} are around an order of a magnitude smaller in terms of total number of scans, and contain 2 to 4 times less annotated scans.
    The sensor was mounted on a static platform with limited background variability at a height of \SI{80}{\cm} where individual legs are usually not visible.

    Our dataset shows interesting interactions between persons and their walking aids with varying backgrounds and a lot of real-world clutter, making it an interesting and challenging dataset.

    \section{METHOD AND EXTENSIONS}
    In order to give an overview of our proposed extensions, we first summarize the relevant parts of the DROW method.
    We refer the interested reader to~\cite{Beyer16RAL} for a more detailed treatment.
    We directly show and discuss resulting precision-recall curves in the following sections, more details of our experimental setup are provided in Section~\ref{sec:experiments}.

    \subsection{DROW Baseline}
    The DROW detector, given a laser scan $s$  will first ``cut out'' a window of fixed real-world size around each individual point $i$: $w_i = \text{cut}(s,i)$.
    The opening-angle used by \emph{cut} to determine which laser points are part of such a window of width $\ell$ at distance $s_i$ is computed as $2\tan^{-1}(\frac{\ell}{2s_i})$.
    The \emph{cut} operation additionally involves clamping of fore- and background as well as re-sampling the data in the window to a fixed number of points.
    Second, each \emph{cutout} $w_i$ is then fed to a CNN which predicts
    1) whether that cutout belongs to an object of interest and, if so, 2) a vote regarding the location of the object's center relative to the cutout's location.
    Finally, all votes are collected and cast into a grid centered around the sensor origin, which is then smoothed before finding the maxima and reporting these as detections.
    The precision-recall curves of our re-implementation are shown as dotted lines (\forceref{black,dotted}) in Figure~\ref{fig:method_extensions}.

    \subsection{Voting}\label{sec:voting}
    In~\cite{Beyer16RAL}, votes of all classes were cast into an ``agnostic'' voting grid based on the agnostic \emph{object}-probability $p(O{\mid}w) = \sum_{c \in \mathcal{C}} p(c{\mid}w_i)$ surpassing the detection threshold, and additionally into a class-specific grid according to their class.
    After smoothing and non-maxium suppression (NMS), the cells which correspond to maxima in the agnostic grid generate detections at their center using the class whose class-specific grid has the largest value in the corresponding cell.
    This has several drawbacks; for one, it discretizes the space of generated detections since only grid-centers are reported, and it does not assign class-distributions to detections, which are useful for other components down the line.

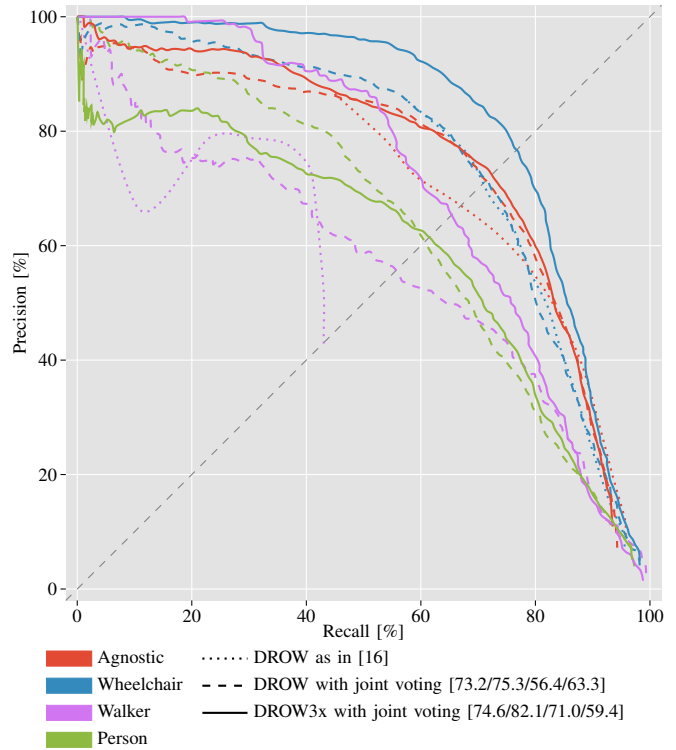
\begin{figure}[t]
        \centering
        \begin{tikzpicture}[]
        \begin{axis}[precrec]
        \draw [gray,dashed] (rel axis cs:0,0) -- (rel axis cs:1,1);
        \precrec{WNet-noP_baseline_wd}{wd_c}{smooth, thick, dotted}
        \precrec{WNet-noP_baseline_wc}{wc_c}{smooth, thick, dotted}
        \precrec{WNet-noP_baseline_wa}{wa_c}{smooth, thick, dotted}

        \precrec{WNet-yesP_wd}{wd_c}{smooth, thick, dashed}
        \precrec{WNet-yesP_wc}{wc_c}{smooth, thick, dashed}
        \precrec{WNet-yesP_wa}{wa_c}{smooth, thick, dashed}
        \precrec{WNet-yesP_wp}{wp_c}{smooth, thick, dashed}

        \precrec{WNet3x_wd}{wd_c}{smooth, thick}
        \precrec{WNet3x_wc}{wc_c}{smooth, thick}
        \precrec{WNet3x_wa}{wa_c}{smooth, thick}
        \precrec{WNet3x_wp}{wp_c}{smooth, thick}
        \end{axis}
        \end{tikzpicture}
        \begin{tikzpicture}
        \begin{customlegend}[
            every axis legend/.append style={nodes={right}},
            legend columns=2,
            legend style={draw=none,font=\scriptsize},
            /tikz/column 2/.style={column sep=5pt},
            legend entries={%
                Agnostic, DROW as in~\cite{Beyer16RAL},
                Wheelchair, DROW with joint voting [\auc{WNet-yesP_wd}/\auc{WNet-yesP_wc}/\auc{WNet-yesP_wa}/\auc{WNet-yesP_wp}],
                Walker, DROW3x with joint voting [\auc{WNet3x_wd}/\auc{WNet3x_wc}/\auc{WNet3x_wa}/\auc{WNet3x_wp}],
                Person,~}]
        \addlegendimage{fill=wd_c, wd_c, sharp plot,area legend}
        \addlegendimage{dotted, sharp plot, thick}
        \addlegendimage{fill=wc_c, wc_c, sharp plot,area legend}
        \addlegendimage{dashed, sharp plot, thick}
        \addlegendimage{fill=wa_c, wa_c, sharp plot,area legend}
        \addlegendimage{sharp plot, thick}
        \addlegendimage{fill=wp_c, wp_c, sharp plot,area legend}
        \end{customlegend}
        \end{tikzpicture}
        \caption{%
            Precision-recall curves showing the impact of our extensions.
            The numbers in brackets correspond to the AUC of the agnostic, wheelchair, walker, and person curves.
            These values are not computable for DROW, see discussion in Section~\ref{sec:voting}.
        }
        \label{fig:method_extensions}
        \vspace{-10pt}
    \end{figure}

    \PAR{Joint voting.} We propose an improved scheme which solves both these problems.
    We still accumulate all the votes into an agnostic voting grid using $p(O{\mid}w)$ and perform smoothing and NMS.
    However, we then assign each individual vote to the maximum in the agnostic grid which is closest to that vote, discarding spurious votes whose distance to the nearest maximum surpasses a certain radius.
    We then report a detection through the mean of its assigned votes, both for the location and the class distribution.
    This results in fully continuous detections as well as well-behaved class distributions.
    For evaluation purposes we simply report the probability of the dominant class, but in practice components processing the detections could make use of the full distribution.
    The precision-recall curves of this new voting scheme (now also including persons as a class) are represented by dashed lines (\forceref{black,dashed}) in Figure~\ref{fig:method_extensions}, showing a major improvement for walkers with wheelchairs being largely unaffected.

    \begin{figure*}
        \centering
        \begin{subfigure}{0.35\textwidth}
            \includegraphics[width=\linewidth]{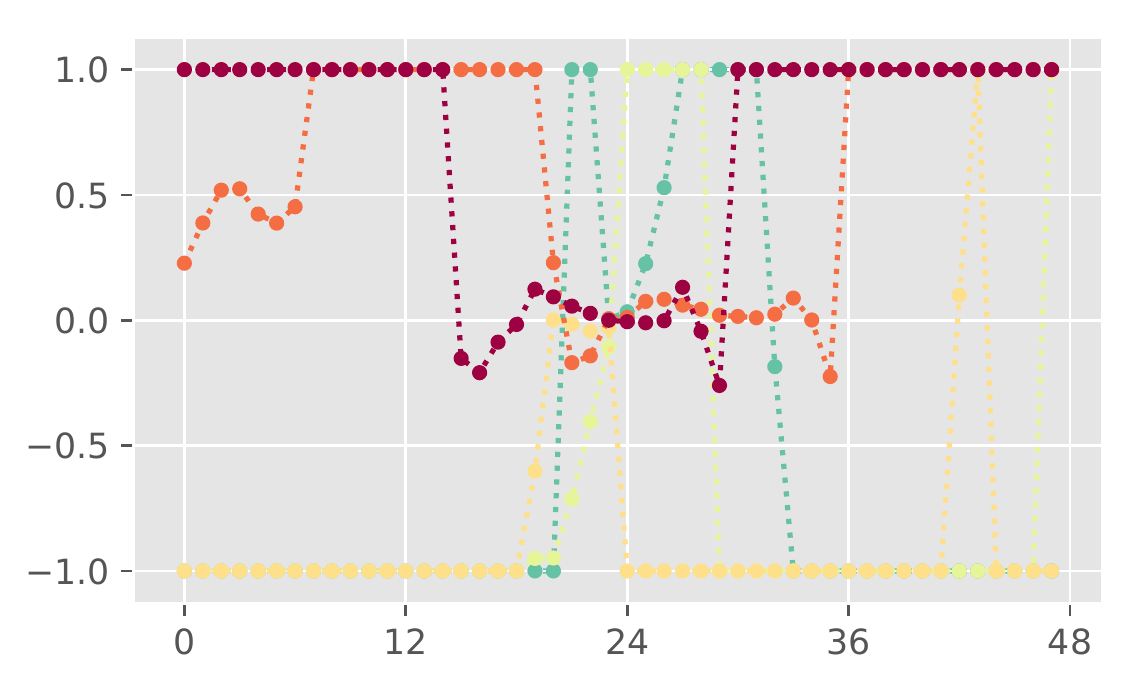}
            \vspace{-20pt}
            \caption{Naive temporal cutout.}
            \label{fig:cutout1}
        \end{subfigure}%
        \begin{subfigure}{0.32\textwidth}
            \includegraphics[width=\linewidth]{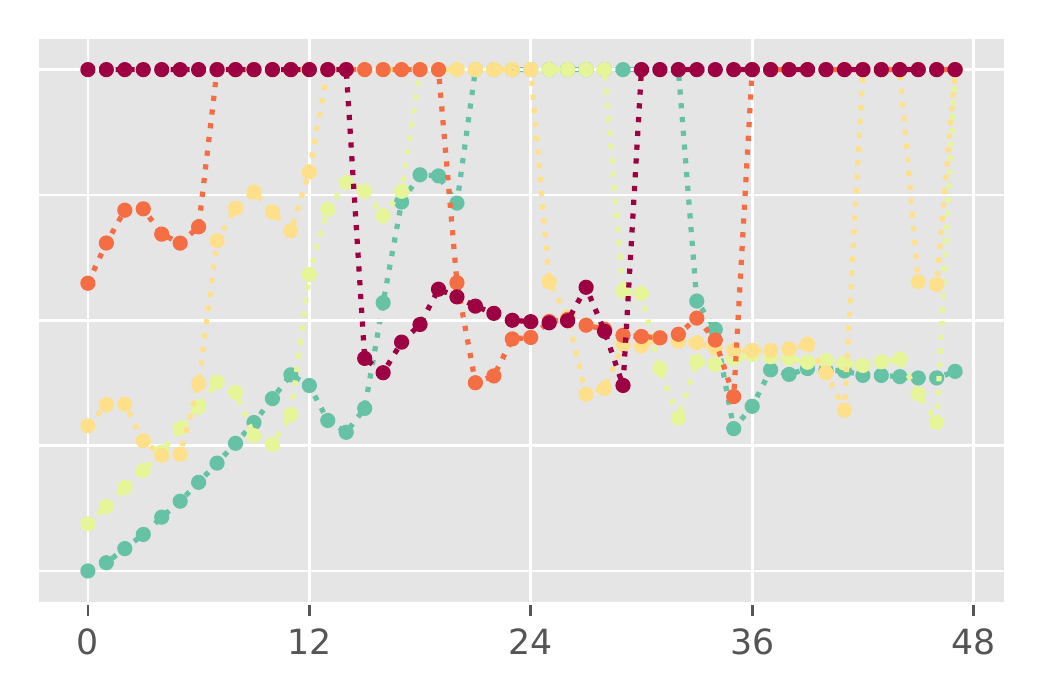}
            \vspace{-20pt}
            \caption{Temporal cutout with fixed location.}
            \label{fig:cutout2}
        \end{subfigure}%
        \begin{subfigure}{0.32\textwidth}
            \includegraphics[width=\linewidth]{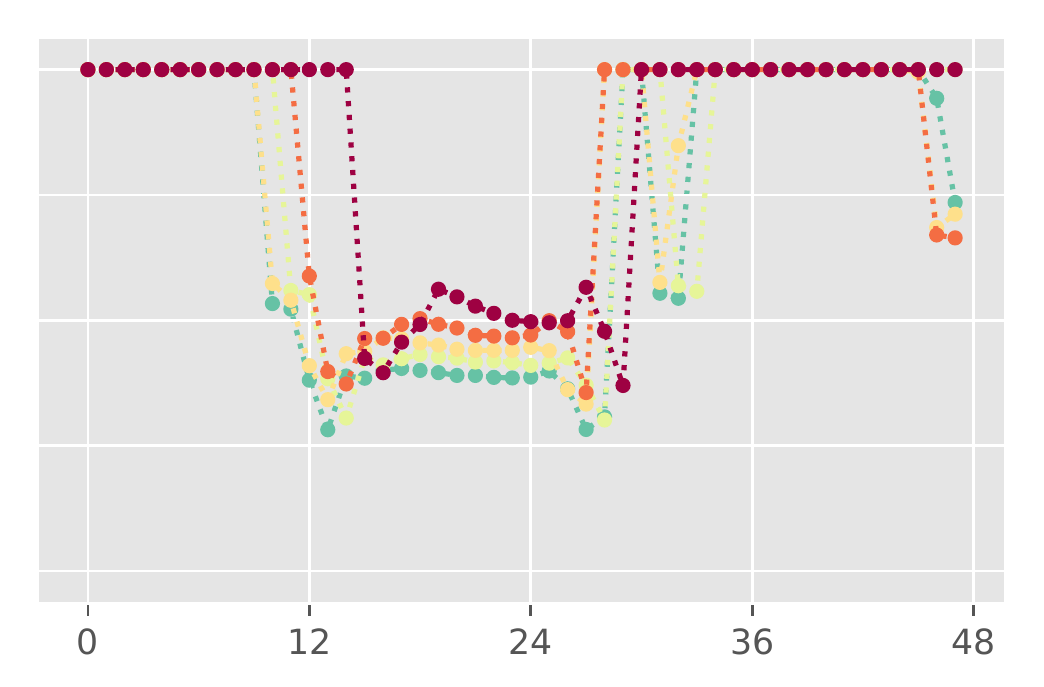}
            \vspace{-20pt}
            \caption{Fixed location and odometry.}
            \label{fig:cutout3}
        \end{subfigure}
        \vspace{-1pt}
        \caption{%
            Variants of temporal cutouts described in Section~\ref{sec:temporal-integration}.
            All three cutouts correspond to the same point $i$ in a scan at time $t$.
            All dots correspond to actual values of $W_i^t$, the color gradient moves from red at $t$ into the past.
            This is the actual input seen by the CNN for that specific point $i$.
        }
        \label{fig:odom_cut_outs}
        \vspace{-10pt}
    \end{figure*}

    \PAR{Split voting.} We also experimented with an alternative voting mechanism where we did not use an agnostic voting grid, but only class-specific ones.
    The advantage here is that it is now possible to apply class-specific blurring.
    However, this then needs another NMS step on the reported detections in order to remove near exact duplicates of different classes.
    Split voting consistently outperformed joint voting by about $1\%$, however, it has multiple additional hyperparameters which severely impact performance.
    Since this method also is a little slower and we aim for a robust detector, we decided to use joint voting instead.

    Knowing which points in the scan contributed to one detection has further advantages, one could for example estimate a confidence for the localization based on how the votes are distributed, or one could determine the spatial extent of the detected person.
    However, we leave these extensions to future work.

    \subsection{Network Architecture}
    We also update the network architecture using current best-practice, the main difference being that it is larger.
    This significantly improves the results, especially for the most difficult walker class, \cf~the solid lines (\forceref{black}) in Figure~\ref{fig:method_extensions}.
    The new network architecture, dubbed ``DROW3x'', consists of three stages of three zero-padded convolutional layers of filter size three each, initialized as in~\cite{He15ICCV}.
    Batch-normalization~\cite{Ioffe15ICML} and Leaky-ReLU~\cite{Maas13ICML} (leak $0.1$) nonlinearities are applied after all convolution operations.
    Each stage is followed by a max-pooling operation of size two and standard dropout~\cite{Srivastava14JMLR}.
    A fourth, final stage consists of two more convolutional blocks as above followed by global average pooling, similarly to~\cite{Szegedy15CVPR,He16ECCV}, after which a single fully-connected layer computes the outputs, \ie SoftMax probabilities and regressed votes.
    We have also experimented with various ResNet~\cite{He16ECCV} and ResNeXt~\cite{Xie17CVPR}-like architectures, but they did not lead to any further improvements while being significantly slower\footnote{They did minimize the loss much better, but this did not translate to gains in precision-recall curves.}.

    \section{TEMPORAL INTEGRATION}
    \label{sec:temporal-integration}
    Another important cue for making sense of the low amount of information conveyed by a single laser scan is the change over time.
    The manual annotation, especially of persons, was only feasibly by scrolling through time and ``tracking'' each person individually in one's head.
    The way the DROW dataset was annotated allows for interesting ways of incorporating the time dimension into a method, although that was not taken advantage of in~\cite{Beyer16RAL}.
    Here, we propose a few simple ways in which that temporal information can be used to great advantage, and we hope that these initial results will spur research into more elaborate ways of using time in detection.
    The general idea is to give the neural network a short snippet of the past in addition to the current measurement.
    Note that this is significantly different from and, in fact, orthogonal to tracking; the resulting method is still a detector.
    In order to do so in the context of the DROW detector, we need to decide on two things:
    1) What is a ``short snippet of windows,'' and 2) where in the neural network is the time dimension fused?

    \PAR{Temporal cutouts.}
    When generating the cutout $w_i^t$ for a given point $i$ of the scan at the current time $t$,
    one could use the cutouts for that same point at the previous times and collect these into a \emph{temporal} cutout
    \begin{equation}\label{eq:time1}
        W_i^t = \Big[\text{cut}\left(s^t, i\right) \cdots \text{cut}\left(s^{t-T}, i\right)\Big].
    \end{equation}
    This naive approach, however, is flawed: if the point $i$ falls onto a leg at times $t$ and $t-2$, but between the legs onto the background at time $t-1$, this approach would lead to the second cutout being centered on whatever is behind that person and thus giving arbitrary, wrong temporal context information as shown in Figure~\ref{fig:cutout1}.
    Instead, it is more meaningful to fix a cutout's location $r$ according to the current time $t$, and use that fixed location for cutting out the same window at the previous times as temporal context:
    \begin{equation}\label{eq:time2}
        W_i^t = \Big[\text{cut}\left(s^t, i, r^t\right) \cdots \text{cut}\left(s^{t-T}, i, r^{t-T}\right)\Big].
    \end{equation}
    This results in much more consistent windows, see also Figure~\ref{fig:cutout2}.
    However, more difficulties keeping the window fixed in real-world location arise when the robot turns:
    when turning, a point at index $i$ in the scan at time $t$ will be in another index $j$ at time $t-1$.
    This difference $\Delta = i - j$ can be provided by most mobile platform's hardware through the odometry, and can be taken into account for the temporal cutouts as follows:
    \begin{equation}\label{eq:time3}
        W_i^t = \Big[\text{cut}\left(s^t, i, r^t\right) \cdots \text{cut}\left(s^{t-T}, i - \Delta^{t-T}, r^{t-T}\right)\Big].
    \end{equation}
    The final, stabilized cutout is shown in Figure~\ref{fig:cutout3}.
    A further possible improvement would be to make use of the translational component of the odometry, however, translating a laser scan correctly is an ill-posed problem that we do not address in this work.

    \PAR{Temporal fusion.}
    Now that we have a way to generate a meaningful temporal cutout $W_i^t$ for each individual point $i$ in the current scan $s^t$,
    the next question is how this temporal cutout is most effectively used by the neural network.
    The most obvious choice would be to simply stack the entries in the \emph{channel} dimension of the network's input; this is also known as \emph{early fusion} (EF).
    As has been shown most prominently by~\cite{Mnih13NIPSW}, this is already a reasonable approach that allows the network to learn something about motion in the scene.
    Further research~\cite{Feichtenhofer16CVPR} has shown that fusing temporal information later into the network, \ie \emph{late fusion} (LF) can be more effective.
    We achieve this by replicating the first two blocks of the network in the time dimension, keeping the weights tied, and passing their sum as input to the third block of the network.
    This way, the network keeps the exact same amount of parameters, allowing for a direct comparison of the impact of time in Table~\ref{table:temporal_fusion}.
    It is likely that more extensive study of the fusion mechanism, including the use of recurrent neural networks (RNNs), will lead to even better results.
    This is not the major focus of our work and left for future work.

\begin{figure}[t]
        \centering
        \begin{tikzpicture}[]
        \begin{axis}[precrec]
        \draw [gray,dashed] (rel axis cs:0,0) -- (rel axis cs:1,1);
        \precrec{WNet3x_wd}{wd_c}{smooth, thick, dotted}
        \precrec{WNet3x_wc}{wc_c}{smooth, thick, dotted}
        \precrec{WNet3x_wa}{wa_c}{smooth, thick, dotted}
        \precrec{WNet3x_wp}{wp_c}{smooth, thick, dotted}

        \precrec{WNet3xLF2p-T5-odom=False-center=each_wd}{wd_c}{smooth, thick, densely dotted}
        \precrec{WNet3xLF2p-T5-odom=False-center=each_wc}{wc_c}{smooth, thick, densely dotted}
        \precrec{WNet3xLF2p-T5-odom=False-center=each_wa}{wa_c}{smooth, thick, densely dotted}
        \precrec{WNet3xLF2p-T5-odom=False-center=each_wp}{wp_c}{smooth, thick, densely dotted}

        \precrec{WNet3xLF2p-T5-odom=False_wd}{wd_c}{smooth, thick, dashed}
        \precrec{WNet3xLF2p-T5-odom=False_wc}{wc_c}{smooth, thick, dashed}
        \precrec{WNet3xLF2p-T5-odom=False_wa}{wa_c}{smooth, thick, dashed}
        \precrec{WNet3xLF2p-T5-odom=False_wp}{wp_c}{smooth, thick, dashed}

        \precrec{WNet3xLF2p-T5-odom=rot_wd}{wd_c}{smooth, thick}
        \precrec{WNet3xLF2p-T5-odom=rot_wc}{wc_c}{smooth, thick}
        \precrec{WNet3xLF2p-T5-odom=rot_wa}{wa_c}{smooth, thick}
        \precrec{WNet3xLF2p-T5-odom=rot_wp}{wp_c}{smooth, thick}
        \end{axis}
        \end{tikzpicture}
        \begin{tikzpicture}
        \begin{customlegend}[
            every axis legend/.append style={nodes={right}},
            legend columns=2,
            legend style={draw=none,font=\scriptsize},
            /tikz/column 2/.style={column sep=5pt},
            legend entries={
                Agnostic, No time [\auc{WNet3x_wd}/\auc{WNet3x_wc}/\auc{WNet3x_wa}/\auc{WNet3x_wp}],
                Wheelchair, Naive late-fusion (Eq.~\ref{eq:time1}) [\auc{WNet3xLF2p-T5-odom=False-center=each_wd}/\auc{WNet3xLF2p-T5-odom=False-center=each_wc}/\auc{WNet3xLF2p-T5-odom=False-center=each_wa}/\auc{WNet3xLF2p-T5-odom=False-center=each_wp}],
                Walker, + Fixed location (Eq.~\ref{eq:time2}) [\auc{WNet3xLF2p-T5-odom=False_wd}/\auc{WNet3xLF2p-T5-odom=False_wc}/\auc{WNet3xLF2p-T5-odom=False_wa}/\auc{WNet3xLF2p-T5-odom=False_wp}],
                Person, + Odometry-correction (Eq.~\ref{eq:time3}) [\auc{WNet3xLF2p-T5-odom=rot_wd}/\auc{WNet3xLF2p-T5-odom=rot_wc}/\auc{WNet3xLF2p-T5-odom=rot_wa}/\auc{WNet3xLF2p-T5-odom=rot_wp}]}]
        \addlegendimage{fill=wd_c, wd_c, sharp plot,area legend}
        \addlegendimage{dotted, sharp plot, thick}
        \addlegendimage{fill=wc_c, wc_c, sharp plot,area legend}
        \addlegendimage{densely dotted, sharp plot, thick}
        \addlegendimage{fill=wa_c, wa_c, sharp plot,area legend}
        \addlegendimage{dashed, sharp plot, thick}
        \addlegendimage{fill=wp_c, wp_c, sharp plot,area legend}
        \addlegendimage{sharp plot, thick}
        \end{customlegend}
        \end{tikzpicture}
        \caption{%
            Different ways of incorporating temporal context into DROW3x, as described in Section~\ref{sec:temporal-integration}.
            The numbers in brackets correspond to the AUC of the agnostic, wheelchair, walker, and person curves.
        }
        \label{fig:temporal_integration}
        \vspace{-10pt}
    \end{figure}
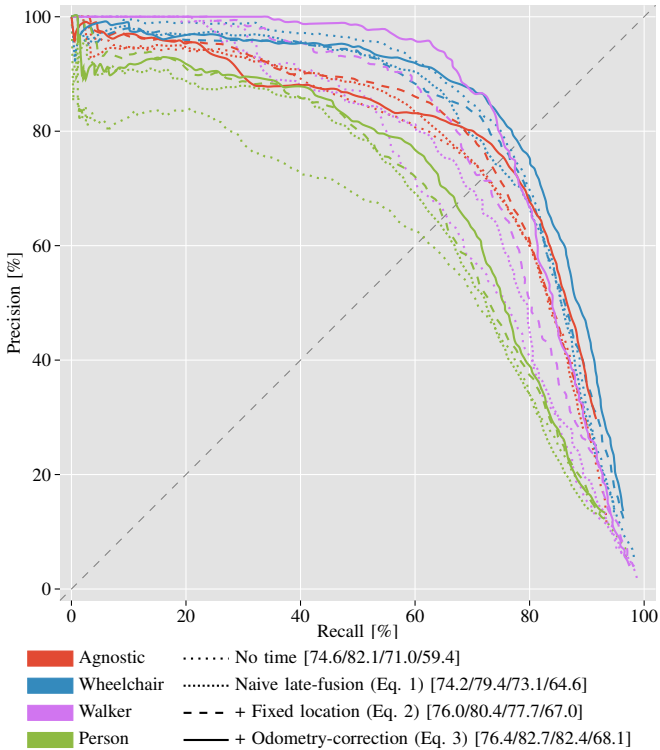

    Figure~\ref{fig:temporal_integration} shows the progression of DROW3x with joint voting from not using time at all (\forceref{black,dotted}) to the naive temporal cutouts (Eq.~\ref{eq:time1}~\forceref{black,densely dotted}), the fixed-location cutouts (Eq.~\ref{eq:time2}~\forceref{black, dashed}), finally leading to the fully ``corrected'' temporal cutouts (Eq.~\ref{eq:time3}~\forceref{black}).
    The main observation here is that while using time is \emph{always} better than not using it, making the input to the neural network as consistent as possible helps even more.

    \section{BASELINES}
    \label{sec:baselines}
    We compare our proposed method against three state-of-the-art methods for person detection in 2D laser described in the following subsections.
    The segment-based person detector by Arras \etal~\cite{Arras07ICRA}, the ROS leg detector~\cite{Pantofaru10ROS} and the joint leg tracker by Leigh \etal \cite{Leigh15ICRA}.
    Figure~\ref{fig:baselines} shows precision-recall curves for all baselines and our proposed approach (\rf{plot:drow3xlf}) on the task of person detection.
    For each evaluated method, we also report percentage values for area under the curve (AUC), equal-error rate (EER) and peak-F1 measure.

    \newcolumntype{C}{>{\centering\arraybackslash}X}
    \newcommand{\fusionline}[2]{#1 & \auc{#2_wc}&\peakfone{#2_wc}&\eer{#2_wc} && \auc{#2_wa}&\peakfone{#2_wa}&\eer{#2_wa} && \auc{#2_wp}&\peakfone{#2_wp}&\eer{#2_wp}\\}
    \begin{table}[t]

    \caption{Comparing temporal fusion strategies.}
    \label{table:temporal_fusion}
    \setlength{\tabcolsep}{0pt}
    \setlength{\extrarowheight}{5pt}
    \renewcommand{\arraystretch}{0.75}
    \begin{tabularx}{\linewidth}{p{2.4cm}CCCp{7pt}CCCp{7pt}CCC}

        \toprule
        & \multicolumn{3}{c}{wheelchair} &&  \multicolumn{3}{c}{walker} &&   \multicolumn{3}{c}{person} \\
        \cmidrule[0.5pt]{2-4} \cmidrule[0.5pt]{6-8} \cmidrule[0.5pt]{10-12}
        & AUC & p-F1 & EER && AUC & p-F1 & EER && AUC & p-F1 & EER \\
        \midrule
        \fusionline{No time}{WNet3x}
        \midrule
        \fusionline{Naive Early-fusion}{WNet3xEF-T5-odom=False-center=each}
        \fusionline{+ Fixed location}{WNet3xEF-T5-odom=False}
        \fusionline{+ Odom correction}{WNet3xEF-T5-odom=rot}
        \midrule
        \fusionline{Naive Late-fusion}{WNet3xLF2p-T5-odom=False-center=each}
        \fusionline{+ Fixed location}{WNet3xLF2p-T5-odom=False}
        + Odom correction & \textbf{\auc{WNet3xLF2p-T5-odom=rot_wc}}&\textbf{\peakfone{WNet3xLF2p-T5-odom=rot_wc}}&\textbf{\eer{WNet3xLF2p-T5-odom=rot_wc}} && \textbf{\auc{WNet3xLF2p-T5-odom=rot_wa}}&\textbf{\peakfone{WNet3xLF2p-T5-odom=rot_wa}}&\textbf{\eer{WNet3xLF2p-T5-odom=rot_wa}} && \textbf{\auc{WNet3xLF2p-T5-odom=rot_wp}}&\textbf{\peakfone{WNet3xLF2p-T5-odom=rot_wp}}&\textbf{\eer{WNet3xLF2p-T5-odom=rot_wp}}\\
        \bottomrule
    \end{tabularx}
    \end{table}

    For this evaluation, we do not compare against the Gandalf detector~\cite{Weinrich14ROMAN}, which was already part of the comparison of the original DROW paper~\cite{Beyer16RAL}; while it is one of the few multi-class detectors for 2D range data, it is not straight forward to evaluate for person detection since the available code often produces leg detections, rather than person centroids, and requires an additional data association / tracking step to merge leg detections.
    Like the training code, this has not been made publicly available.
    Nevertheless, our DROW baseline outperformed the Gandalf detector in~\cite{Beyer16RAL}.
    Finally, the resulting detections do not have a confidences, which means the evaluation would result in single precision-recall points.

    \subsection{Segment-based person detector by Arras et al.}\label{sec:arras}
    The original approach for person detection in 2D range data by Arras \etal~\cite{Arras07ICRA}, which forms the basis of the other two methods described below, combines a jump distance segmentation of the laser scan with a boosted classifier trained on hand-crafted geometric features.
    The jump distance segmentation threshold is configured such that two legs typically end up in the same segment.

    We use our ROS-based reimplementation \cite{Linder16ICRA}, in which the Adaboost classifier has been replaced by a better-performing random forest.
    We retrain the detector on the person annotations of the DROW training set, as the model from~\cite{Linder16ICRA} assumed a different sensor mounting height of 70--80 cm where legs are often not visible.
    Automated hyperparameter tuning (Sec.~\ref{sec:experiments}) results in a segmentation threshold of \SI{23}{cm} and a random forest with 128 trees of max depth 50.
    During training, a segment is regarded as foreground if over half of its points end up in a circle of \SI{25}{cm} radius around the annotated person centroid, and as background if over half of the points in a segment do not fall into such a circle.
    All remaining segments are considered as ambiguous and not used for training. The resulting precision-recall curve is shown in Figure~\ref{fig:baselines} (\rf{plot:arras_retrained}).

    As we observed during initial experiments that false positive person detections often occur at the left/right boundaries of the sensor field of view, we incorporate an additional geometric feature that measures the angular difference of the segment center to the boundary of the sensor FOV, which improves performance in AUC by 2.6\% (\rf{plot:arras_retrained_boundary_feat}).

    \subsection{ROS leg detector}
    \label{sec:rosleg}
    This ``detector''~\cite{Pantofaru10ROS} (\rf{plot:ros_leg_detector}) is often used for experiments in robotics as it is readily available as a standard ROS package with a pre-trained model.
    It builds on top of the geometric features from~\cite{Arras07ICRA}, but applies them to individual leg clusters using a random forest classifier.
    The detected legs are then associated and \emph{tracked} over time using a Kalman filter-based tracker.
    For our evaluation, we treat the resulting tracks on a frame-by-frame level as if they were detections.
    Following the findings in~\cite{Leigh15ICRA}, we reduce the leg reliability threshold from 0.7 to 0.3 as otherwise only very few person tracks are initiated.

    \subsection{Joint leg tracker by Leigh et al.}
    \label{sec:leigh}
    The joint leg tracker by Leigh \etal \cite{Leigh15ICRA} is an improved and extended variant of \cite{Pantofaru10ROS} which jointly tracks two legs by allowing up to two associations of detected leg clusters with a person track.
    Similarly to the ROS leg detector, because our evaluation is based upon person centroids as opposed to individual legs, we treat the output of their tracker as detections by considering only non-occluded, matched tracks which have been associated with at least one leg detection in the current frame\footnote{Preliminary experiments showed that inclusion of occluded tracks leads to AUC/EER scores that are worse by around 4\%.}.
    Detection confidences are computed from track confidence scores.
    We alter the original training method, which assumes that independent data recordings with positive and negative training examples are available, by assigning person annotations to potential leg clusters within a given radius, \eg 25 cm.
    For the purpose of a fair evaluation, we disable their proposed online occupancy grid mapping extension\footnote{We were also not able to run it reliably on all our sequences.}, which is used to suppress false positive detections in walls and other static structures;
    however, \emph{all} evaluated methods could similarly be extended to suppress false detections \eg by using a static map of the environment as shown in~\cite{Linder16ICRA}.

\begin{figure}[t]
        \centering
        \begin{tikzpicture}[]
        \begin{axis}[precrec]
        \draw [gray,dashed] (rel axis cs:0,0) -- (rel axis cs:1,1);
        \precrec{ros_leg_detector}{good_gray}{smooth, thick}\label{plot:ros_leg_detector}
        \precrec{ros_leg_detector_0.3}{good_gray}{smooth, dotted}\label{plot:ros_leg_detector_r}
        \precrec{arras_retrained}{deep_pink}{smooth, thick}\label{plot:arras_retrained}
        \precrec{arras_retrained_0.3}{deep_pink}{smooth, dotted}\label{plot:arras_retrained_r}
        \precrec{arras_retrained_with_boundary_feat_1k}{random_blue}{smooth, thick}\label{plot:arras_retrained_boundary_feat}
        \precrec{arras_retrained_with_boundary_feat_1k_0.3}{random_blue}{smooth, dotted}\label{plot:arras_retrained_boundary_feat_r}
        \precrec{leigh_untrained}{awesome_orange}{smooth, thick}\label{plot:leigh_untrained}
        \precrec{leigh_untrained_0.3}{awesome_orange}{smooth, dotted}\label{plot:leigh_untrained_r}
        \precrec{leigh_retrained_hyperopted}{shit_brown}{smooth, thick}\label{plot:leigh_retrained}
        \precrec{leigh_retrained_hyperopted_0.3}{shit_brown}{smooth, dotted}\label{plot:leigh_retrained_r}
        \precrec{WNet3xLF2p-T5-odom=rot_wp}{wp_c}{smooth, thick}\label{plot:drow3xlf}
        \precrec{WNet3xLF2p-T5-odom=rot_wp_0.3}{wp_c}{smooth, dotted}\label{plot:drow3xlf_r}
        \precrec{WNet3xLF2p-T5-odom=rot_wd}{wd_c}{smooth, thick}\label{plot:drow3xlf_agn}
        \precrec{WNet3xLF2p-T5-odom=rot_wd_0.3}{wd_c}{smooth, dotted}\label{plot:drow3xlf_agn_r}
        \precrec{WNet3xLF2p-T5-odom=rot-trainval_wp}{evil_green}{smooth, thick}\label{plot:drow3xlf_val}
        \precrec{WNet3xLF2p-T5-odom=rot-trainval_wp_0.3}{evil_green}{smooth, dotted}\label{plot:drow3xlf_val_r}
        \end{axis}
        \end{tikzpicture}
        \vspace{-8pt}

        \begin{tikzpicture}[]
        \begin{customlegend}[
            every axis legend/.append style={nodes={right}},
            legend columns=2,
            legend style={draw=none,font=\small},
            /tikz/column 2/.style={column sep=5pt},
            legend entries={
                ~\\
                AUC\%~~~peak F1\%~~~EER\%\\
                ROS leg detector\\\aucpeakfoneeer{ros_leg_detector}\\
                Arras retrained\\\aucpeakfoneeer{arras_retrained}\\
                Arras retrained + feat\\\aucpeakfoneeer{arras_retrained_with_boundary_feat_1k}\\
                Leigh off-the-shelf\\\aucpeakfoneeer{leigh_untrained}\\
                Leigh retrained\\\aucpeakfoneeer{leigh_retrained_hyperopted}\\
                DROW3x \\\aucpeakfoneeerbbb{WNet3xLF2p-T5-odom=rot_wp}\\
                DROW3x (agnostic)\\\aucpeakfoneeeriii{WNet3xLF2p-T5-odom=rot_wd}\\
                DROW3x + validation\\\aucpeakfoneeeriii{WNet3xLF2p-T5-odom=rot-trainval_wp}\\}]
        \bl \bl
        \addlegendimage{sharp plot, good_gray, ultra thick} \bl
        \addlegendimage{sharp plot, deep_pink, ultra thick} \bl
        \addlegendimage{sharp plot, random_blue, ultra thick} \bl
        \addlegendimage{sharp plot, awesome_orange, ultra thick} \bl
        \addlegendimage{sharp plot, shit_brown, ultra thick} \bl
        \addlegendimage{sharp plot, wp_c, ultra thick} \bl
        \addlegendimage{sharp plot, wd_c, ultra thick} \bl
        \addlegendimage{sharp plot, evil_green, ultra thick} \bl
        \end{customlegend}
        \end{tikzpicture}
        \caption{
            Different baselines and our proposed approach (DROW3x) evaluated on the person class of our test set.
            The thin dotted lines show the same experiment, using an evaluation radius of \SI{0.3}{\m}.
            Unlike all other curves the red line represents an agnostic evaluation, meaning we use detections and annotations of all classes jointly.
        }
        \label{fig:baselines}
        \vspace{-10pt}
    \end{figure}
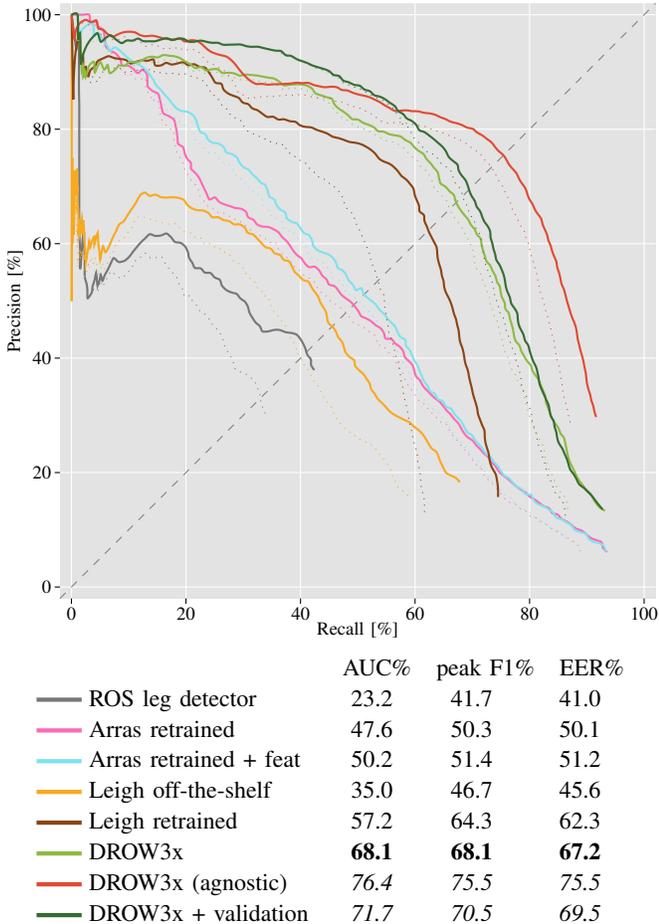

    In one experiment, \cf Figure~\ref{fig:baselines} (\rf{plot:leigh_untrained}), we use the provided model off-the-shelf with default hyperparameters, and see that it performs slightly better than the off-the-shelf ROS leg detector, but worse than the retrained Arras detector. In another experiment, we retrain the classifier and extensively tune 10 different hyperparameters, including leg association and track initiation/continuation thresholds, segmentation distance, confidence update factor as well as class weighting and person radius during training. This leads to around +20\% better results (\rf{plot:leigh_retrained}).
    The large difference in performance can partly be explained by the dependence of the learned model's features on the angular resolution of the sensor (originally 1/3 degree).


    We can also see that the joint leg tracker, even after extensive automatic hyperparameter optimization, achieves a maximum recall of only around 75\%, while both our proposed approach and the Arras detector can achieve up to 95\% recall.
    Further experiments show that the method of Leigh outputs basically no person detections beyond a distance of around 7m
    , whereas both DROW3x and the detector by Arras still generate detections up to the maximum detection distance of 15m.
    The reason for this appears to be that at 7m distance and an angular resolution of 1/2 degree, a single leg cluster often consists of only two or less points, while some of the used geometric features by design require at least three points\footnote{We tried to set the values of affected features to a constant in such cases and retrain the classifier, but neither achieved better results, nor real-time performance ($<1$ Hz) using a lower minimum point count per leg cluster.}.
    Our implementation of the Arras detector, which uses a similar set of geometric features, does not suffer from this issue because the entire person (including both legs) is considered as one segment upon which features are computed.
    We believe the limited maximum range is a significant limitation of the leg-tracking method of \cite{Leigh15ICRA}, as many robots are still equipped with 2D laser range finders with an angular resolution of 1/2 deg or less (\eg \cite{Beyer16RAL,Weinrich14ROMAN,Luber11IJRR,Spinello08ICRA}), especially when considering the increasing adoption of low-cost, low-res 2D laser scanners in domestic applications.


    \section{EXPERIMENTAL SETUP}\label{sec:experiments}
    For the experiments in this paper, we re-implemented our original DROW detector in PyTorch~\cite{Paszke17NIPSWorkshop} from scratch.
    We now use the Adam~\cite{Kingma15ICLR} optimizer and fix a learning rate schedule starting at $10^{-3}$ for $40$ epochs, then exponentially decaying to $10^{-6}$ within 10 epochs.
    Initial comparisons showed this not to affect the scores much, but it stabilizes results at the end of training and makes experiments comparable without the need to cherry-pick.

    Unless stated otherwise, our DROW detector is always trained for three classes (persons, wheelchairs and walkers), however, to avoid cluttered plots we only show the precision-recall curves for persons in Figure~\ref{fig:baselines}. Initial experiments show that no significant improvements were obtained by training a person-only version of our detector.

    We use Hyperopt~\cite{Bergstra13ICML} to tune most hyperparameters of all methods (except neural network training) on the validation set\footnote{We allocate the same computational budget for each experiment.}.
    We optimize the sum of the AUCs of all available precision-recall curves.
    For our DROW baselines, the old voting mechanism does not produce dense precision-recall curves, thus making it difficult to compute AUCs.
    In this case, we optimize the sum of the peak-F1 scores.

    During evaluation a detection is considered correct if the centroid lies within \SI{0.5}{\m} radius of an annotation.
    At most one detection can be assigned to each annotation.

    \section{DISCUSSION}

    Looking at the results in Figure~\ref{fig:baselines}, one can distinguish between three types of methods.

    \PAR{Leg-tracking based}methods which detect individual legs and assemble those to person tracks (Section~\ref{sec:rosleg} and~\ref{sec:leigh}) can, when re-trained on our large dataset, perform quite well.
    Their recall is limited by the methods' design, since individual legs cannot be detected at large distances.
    Furthermore, due to the strong assumptions regarding shape and motion, this type of method has the severe drawback of only being applicable to persons in pants; it does not readily apply to other clothing or different classes, such as wheelchairs.

    \PAR{Segment-classification based}methods segment the scan into individual pieces which are then classified as person or background (Section~\ref{sec:arras}).

    This type of method does not suffer from the aforementioned drawbacks but, even though trained on our dataset, performs significantly worse than a re-trained leg-tracking based method.
    Note that due to learning about persons, as opposed to legs, it can reach very high recall values.

    \PAR{The DROW}methods are based on deeply learned voting, and do neither suffer from the principled drawbacks of the leg-tracking type, nor from the inferior performance of the segment-classification type.
    Specifically, the light green curve (\rf{plot:drow3xlf}) shown in Figure~\ref{fig:baselines} corresponds to the person detection performance of DROW \emph{while it simultaneously detects walkers and wheelchairs}, which none of the other methods can do.
    This already works well with the baseline DROW3x, but the simplistic use of temporal information (without doing any tracking) further improves the results, beating current state-of-the-art by $10.9\%$.
    Additionally, the red curve (\rf{plot:drow3xlf_agn}) shows the agnostic performance, meaning if the only task is to detect people and walking aids jointly, our performance even improves.


    \PAR{Temporal and odometry}information are clearly beneficial when they can be integrated into a method.
    This is the case for DROW, as we have shown in Section~\ref{sec:temporal-integration}, and in initial experiments with the Joint leg tracker, we saw an increase in EER of 6\% when using odometry.
    In~\cite{Arras07ICRA}, Arras~\etal mention that including simple motion features led to worse performance, suggesting that it is not always obvious how to make good use of temporal information.
    We suggest a simple but principled way of using temporal information and odometry in DROW, which shows consistent improvements, but we look forward to more research on this subject.


    \PAR{A larger training set} further improves our results slightly (\rf{plot:drow3xlf_val}).
    For this experiment, we merged the training and validation sets and use the hyperparameters selected on the validation set previously.
    The improved performance shows that the model has additional capacity and indicates that even better detectors can be learned given more data.

    \PAR{The localization accuracy} of the detections can be investigated by evaluating with a smaller radius of \SI{0.3}{\m}.
    The corresponding curves are shown by the dashed lines in Figure~\ref{fig:baselines} (\forceref{black, thin, dotted}).
    All leg-tracking based approaches perform significantly worse under this stricter evaluation scheme.
    A possible explanation for this behavior is the extra latency introduced by the leg trackers' Kalman filter\footnote{It should be noted that the evaluation in~\cite{Leigh15ICRA} used a radius of \SI{0.75}{m}, which is too lenient to be an indicator for practical use.}.
    Meanwhile, both the segment-classification and DROW based approaches are hardly affected.
    This means their detections are located more precisely, which is another clear advantage.

    \PAR{Hyperopt}resulted in surprising parameter values for DROW.
    Considering all the hyperparameters of our method are very interpretable, some final values differed a lot from our initial ``sane'' manually selected ones.
    For DROW the final test performance was a little better when using our initial guesses, indicating that hyperopt might have overfitted the validation set.
    Nevertheless, the Leigh tracker gained about 4\% AUC and, to be consistent, we report hyperopt tuned performances for all DROW curves.
    Since we optimized the AUC of all class curves jointly, especially the person performance typically slightly decreased.
    If only person detections are relevant, we could easily improve the person performance of DROW at the cost of the other classes.
    The fact that DROW still performs very well, even with some unintuitive values, suggests it is robust and can also easily be tuned.


    \PAR{Multi-target tracking systems} are usually used in tandem with detectors in practice~\cite{Hawes17RAM,Triebel16FSR}.
    This facilitates higher-level analysis~\cite{Beyer15GCPR} and reasoning, especially across longer timeframes~\cite{Hawes17RAM}.
    Nevertheless, in~\cite{Linder16ICRA} Linder and Breuers~\etal make the ``observation that [...] detector performance is the single, most important factor influencing tracking performance which goes far beyond the impact of the chosen tracking algorithm.''
    This is why the development of strong, multi-class detectors such as DROW is necessary for mobile robotics.

    \section{CONCLUSIONS}
    For mobile robot platforms that operate among people, it is a crucial ability to detect those people, be it with or without mobility aids.
    These robots are usually equipped with 2D laser scanners for safety reasons, making the ability to piggy-back this modality for detection especially attractive.
    Due to the sparsity of the signal, though, this task is extremely challenging.

    We release a large new set of person annotations recorded ``in the wild'', making this, to the best of our knowledge, the largest publicly available dataset for person detection in 2D range data.
    We evaluate several state-of-the-art detectors as baselines on this dataset, ready for others to compare against. (All precision-recall curves will be provided online.)

    We extend the DROW detector to the task of detecting persons---in addition to, and  simultaneously with wheelchairs and walkers---hereby introducing to the field an interesting departure from the classic segment-and-classify pipeline, outperforming the strongest baseline by $10.3\%$ AUC.
    We solve shortcomings of the voting scheme presented in the original DROW paper, namely the discretization of detections and the inability to report meaningful confidences.

    Inspired by how we as humans annotated the dataset\footnote{Annotation would have been impossible without supporting images and the full-scene understanding of a human brain.}, we propose initial ways of integrating temporal information into DROW, leading to substantial gains in performance, especially for the challenging person and walking aid classes.
    This suggests more research in this area can lead to even better detectors.

    Most importantly, this work shows that DROW is the only strong, unbiased multi-class detector for 2D range data.

    \section*{ACKNOWLEDGMENT}
    We would like to thank Supinya Beyer, without whose help in annotating the data this paper would not have been possible, and Narunas Vascevicius for helpful discussion.
    This work has been funded, in part, by the EU Horizon 2020 projects ILIAD (H2020-ICT-2016-732737) and CROWDBOT (H2020-ICT-2017-779942), as well as the BMBF project FRAME (16SV7830).

    \bibliographystyle{IEEEtran}
    \bibliography{abbrev_short,bib}
\end{document}